\documentclass{article}

\usepackage{arxiv}

\usepackage[utf8]{inputenc} 
\usepackage[T1]{fontenc}    
\usepackage{hyperref}       
\usepackage{url}            
\usepackage{booktabs}       
\usepackage{amsfonts}       
\usepackage{nicefrac}       
\usepackage{microtype}      
\usepackage{lipsum}
\usepackage{graphicx}
\usepackage{amssymb}
\usepackage{amsmath}
\newcommand{\R}{\rm I\!R}

\usepackage{flushend}
\usepackage{algorithm} 
\usepackage{algpseudocode} 
\usepackage{graphicx, caption, subcaption}
\usepackage{hyperref}
\usepackage{dirtytalk}
\graphicspath{ {./images/} }

\title{A non-alternating graph hashing algorithm for large scale image search}

\author{
 Sobhan Hemati \\
 Kimia Lab\\
 University of Waterloo\\
 ON, Canada \\
\And
Mohammad Hadi Mehdizavareh \\
School of Electrical and Computer Engineering\\
University of Tehran\\
Tehran, Iran  \\
\And
Shojaeddin Chenouri \\
Department of Statistics and Actuarial Science\\
University of waterloo\\
ON, Canada \\
\And
Hamid R Tizhoosh \thanks{Corresponding author}\\
Kimia Lab and Vector Institute, MaRS Center\\
University of Waterloo\\
ON, Canada\\
\texttt{tizhoosh@uwaterloo.ca} 
}

\begin{document}

\maketitle
\begin{abstract}
In the era of big data, methods for improving memory and computational efficiency have become crucial for the successful deployment of technology. Hashing is one of the most effective approaches to deal with the computational limitations associated with big data. One natural way to formulate this problem is spectral hashing, which directly incorporates an affinity to learning binary codes. However, owing to the binary constraints, the optimization becomes intractable. To mitigate this challenge, different relaxation approaches have been proposed to reduce the computational load required to obtain binary codes and still attain a good solution. The problem with all existing relaxation methods involves the use of one or more additional auxiliary variables to attain high-quality binary codes while relaxing the problem. The existence of auxiliary variables leads to the coordinate descent approach, which increases the computational complexity. We argue that the introduction of these variables is unnecessary. To this end, we propose a novel relaxed formulation for spectral hashing that adds no additional variables to the problem. Furthermore, instead of solving the problem in the original space where the number of variables is equal to the data points, we solve the problem in a much smaller space and retrieve the binary codes from this solution. This technique reduces both the memory and computational complexity simultaneously. We apply two optimization techniques, namely, the projected gradient and optimization on the manifold, to obtain the solution. Using comprehensive experiments on four public datasets, we show that the proposed efficient spectral hashing (ESH) algorithm achieves a highly competitive retrieval performance compared with the state-of-the-art algorithms at low complexity.

\end{abstract}

\section{Introduction}
Through the advancement of digital technologies, we are witnessing
a greater explosion in the amount of data generated than ever before. Such data carry valuable information that can help many industries, such as healthcare, achieve more effective procedures. At the same time, as the size of the available data increases, maintaining and extracting value from the data becomes more challenging. Two main issues with big data are the need for significant storage and computational resources. A well-known approach to addressing these challenges is ``hashing,'' which refers to learning similarity preserving binary codes for data, e.g., for images. Learning such representations offers two obvious advantages: First, it considerably reduces the amount of memory required to save massive amounts of data, and second, it improves the computational complexity by the help of Hamming distance. One of the most important use cases of this technique is the approximate nearest neighbor search (ANN), which has many applications. Examples include image annotation \cite{wang2014binary}, large-scale clustering \cite{7298596}, image patch matching \cite{korman2015coherency}, and video segmentation \cite{Liu_2014_CVPR}. In addition, the authors in \cite{hu2018hashing}  showed that the problem of training binary weight networks is equivalent to a hashing problem. As a result, there has recently been increased interest in applying hashing to the network quantization problem. The authors in \cite{gholami2021survey} and \cite{qin2020binary} discussed recent advances in network quantization methods in which hashing algorithms play an important role.

Hashing algorithms can be classified into data-independent and data-dependent methods \cite{wang2017survey}. In data-independent algorithms, no information is used to obtain a binary code. An example is locality sensitivity hashing \cite{4031381} and its extensions \cite{5459466}, where random projections are employed to achieve binary representations. Because there is no training phase with these algorithms, their deployment is easy in practice. However, this simplicity degrades the performance of these algorithms for compact binary codes. Note that the use of high-dimensional binary representations increases both the computational and memory efficiencies, which is contrary to the initial motivations for developing a hashing algorithm.

In data-dependent algorithms, either affinity information or both affinity and label information in the training data are employed to transform the data from the Euclidean space to a compact binary representation in the Hamming space. The former methods are known as unsupervised hashing algorithms (which is the focus of this paper), and the latter are supervised hashing methods. Owing to the incorporation of label information, supervised hashing achieves a superior performance compared to unsupervised cases \cite{Eghbali_2019_CVPR, cakir2017mihash}. Deep learning has
significantly improved the state-of-the-art performance of hashing algorithms \cite{Eghbali_2019_CVPR, cakir2017mihash}. However, in practice, there are two limitations to employing deep hashing algorithms. First, they are usually supervised methods, meaning that they take advantage of both label and affinity representations to obtain binary codes. Second, this significant performance results in high computational and storage overheads. However, in practice, unsupervised hashing may be more useful because in many real-world applications the available data are unlabeled or the data labeling is prohibitively expensive. In addition, owing to the high number of parameters in deep hash functions, their test time encoding can be high, which is contrary to the philosophy behind the use of hashing. 
Although deep learning has mainly been applied to the supervised hashing problem, recent attempts have been made to develop deep unsupervised hashing methods. Some examples include DH \cite{erin2015deep}, UHBDNN \cite{do2016learning}, DeepBit \cite{lin2016learning}, and SADH \cite{shen2018unsupervised}. Another new type of deep hashing method is an architecture based on generative adversarial networks (GANs). For example, HashGAN \cite{cao2018hashgan} and a binary
generative adversarial network (BGAN+) \cite{song2020unified} are representative methods, of which HashGANs require supervision, whereas BGAN+ can be employed under both supervised and unsupervised scenarios.

The main issue in solving hashing problems is the discrete nature of the problem, which leads to difficult optimization problems. In the iterative quantization (ITQ) algorithm \cite{gong2012iterative} the authors proposed a two-step approach to compress and quantize the data. In the first step, the data are projected into a rotation invariance space by applying a PCA. Then, in the next step, using an alternating optimization approach, the optimal rotation is found to minimize the quantization error. A similar two-step approach was employed in isotropic hashing (Isohash) \cite{kong2012isotropic}, where instead of minimizing the quantization loss, the authors equalized the variance across different projections. The K-nearest neighbor hashing algorithm (KNNH) \cite{He_2019_CVPR} is also a new development based on the ITQ skeleton, where a heuristic approach was suggested to minimize the conditional entropy of deriving binary codes. Simultaneous compression and quantization (SCQ) \cite{HOANG2020102852} is another approach based on ITQ that attempts to combine a dimensionality reduction and quantization steps.
Another line of research dealing with challenging optimization problems faced in learning a binary representation is the encoder-decoder architecture. For example, see the binary auto-encoder (BA) \cite{carreira2015hashing}. Furthermore, an alternative for learning similarity-preserving binary codes is the formulation presented in spectral hashing (SH) \cite{weiss2009spectral}, which is based on a graph Laplacian. SH is a natural approach incorporating an affinity matrix directly into
the derived affinity-preserving binary codes. However, the SH method has two main problems. First, calculating the pairwise affinity between data points is computationally expensive, that is, $O(n^2 \times d)$, where $n$ is the number of data points and $d$ is the data dimensionality. Second, even for one bit, the optimization problem in SH is equivalent to balanced graph partitioning, which is a difficult problem. As a result, different attempts have been made to address these limitations.

The focus of this paper is to improve the current solutions for the spectral hashing formulation in terms of both training complexity and retrieval performance. Before presenting our contribution, we briefly review recent developments in spectral hashing.

To reduce the computational complexity of calculating the affinity matrix, the authors of anchor graph hashing (AGH) \cite{liu2011hashing} suggested employing a neighborhood graph to derive a low-rank approximation of the affinity matrix. The complexity of this approximation is $O(n)$. To deal with the computational complexity of the discrete optimization faced in spectral hashing, continuous relaxation was initially proposed. However, in this simplified approach, the binarization step destroys the learned manifold structure and degrades the quality of binary codes. To mitigate this issue, some authors have proposed incorporating a quantization loss. For example, spectral rotation (LGHSR) \cite{li2017large}, a two-step approach similar to the idea in ITQ, was recently introduced to find the optimal rotation that minimizes the quantization loss. In discrete graph hashing (DGH) \cite{liu2014discrete}, the authors attempted to solve the relaxed problem and minimize the distance between the continuous and binary sets simultaneously. In robust discrete hashing (RDSH) \cite{yang2015robust}, an objective function for obtaining a relaxed solution along with both an optimal rotation for reducing the quantization loss and a hash function was developed. Finally, the authors of discrete spectral hashing (DSH) \cite{hu2018discrete} recently proposed another approach to joint optimization over relaxed variables with minimum quantization. They showed that their solution achieves a better performance while reducing the training complexity.

Although different methods have been proposed to achieve better-relaxed solutions while reducing complexity, we should note that all of the proposed methods, including LGHSR, DGH, RDSH, and DSH, employ an alternating approach between two or more optimization variables and keep the binary decision variable within the problem. However, we argue that this is unnecessary.

In this paper, we develop a new formulation that allows us to convert the optimization problem with $n \times k$ parameters to a problem with $d \times k$ decision variables ($d$ dimensionality and $k$ number of bits), where $d\ll n$. In addition, in the proposed formulation, the optimization problem is a function of one decision variable, and as a result, the optimization is non-alternating, which significantly reduces the computational complexity. Two optimization algorithms were employed to obtain a solution for the suggested optimization problem. The results on four public datasets show that the proposed formulation outperforms the state-of-the-art approaches and at the same time is more efficient than the recent algorithms proposed for the graph Laplacian hashing problem.

\section{Spectral Hashing}
Let $\mathbf{X} \in \R^{n\times d}$ denote a mean zero and unit variance data matrix with rows representing training data points, each with a dimensionality of $d$. The original spectral hashing formulation \cite{weiss2009spectral} is as follows: 

\begin{align}\label{eq:1}
 &{\rm argmin}\, \sum_{i,j} A_{i,j} \lVert \mathbf{b}_i -  \mathbf{b}_j \rVert ^2& \\
 &\text{s.t.}\,  \sum_{i} \mathbf{b}_i=\mathbf{0}, \quad \frac{1}{n} \sum_{i} \mathbf{b}_{i} \mathbf{b}^T_{i} = \mathbf{I}_k, \quad  \mathbf{b}_i \in \{-1,1 \}^k & \nonumber
\end{align} 
where $\mathbf{b}_i$ and $\mathbf{b}_j$ are $k$-bit binary codes corresponding to the $i$-th and $j$-th data points, respectively, $\mathbf{I}_k$ is a $k$ by $k$ identity matrix, and $A_{i,j}$ is the $(i,\,j)$ entry of the affinity matrix $\mathbf{A}$, which measures the similarity between the $i$-th and $j$-th data points. Let us denote the $k$-bit binary representation of the data matrix $\mathbf{X}$ by $\mathbf{B} \in \{-1,1 \}^{n \times k}$. Suppose $\mathbf{D}$ is an $n \times n$ diagonal matrix whose diagonal element $i$ is given by $D_{i,i}=\sum_j A_{i,j}$. The matrix form of the optimization problem can then be written as follows:

\begin{align}\label{eq:2}
& \underset{\mathbf{B}}{{\rm argmin}} \quad \textrm{\emph{Tr}}\{\mathbf{B}^{T}(\mathbf{D}-\mathbf{A})\mathbf{B}\}\\
& \text{s.t.} \quad
 \mathbf{B}^T \mathbf{B}=n \mathbf{I}_k, \quad \mathbf{B}^T \mathbf{1}_{n \times 1}=\mathbf{0}, \quad \mathbf{B} \in \{-1,1 \}^{n \times k} \nonumber
\end{align}
where \emph{Tr} represents the trace operation, $\mathbf{L}=\mathbf{D}-\mathbf{A}$ is the graph Laplacian matrix \cite{liu2010large}, and $\mathbf{1}_{n \times 1}$ is a vector of length $n$, where all elements are 1. 

To reduce the computational complexity in calculating the affinity matrix $\mathbf{A}$ from $O(n^2 d)$ to $O(n)$, we employ the low-rank approximation of \cite{liu2011hashing}. This approximation is based on a small number ($m$) of data points, that is, anchors, with $m \ll n$. In general, the $m$ cluster centers obtained from the K-means algorithm, after running a few iterations, are considered as $m$ anchors $u_j$ for $j=1,\,\dots,\, m$. These anchors are used to construct an $n \times m$ affinity matrix $\mathbf{Z}$ using a Gaussian kernel of pairwise distances between the $n$ observed data points and the $m$ anchors. That is, the $(i,\,j)$ entry of the matrix $\mathbf{Z}$ is given by $Z_{i,j}={K(x_i,u_j)}/{N_0}$, where $x_i$ is the $i$-th data point, $K(x_i,u_j)=\exp({\|x_i-u_j\|^2_2}/{\sigma^2})$, and $\sigma$ is a hyperparameter. To impose sparsity on $\mathbf{Z}$, only distances from the $s\ll m$ nearest neighbor anchors are kept, and the rest are set to zero. Furthermore, the normalization factor $N_0$ is calculated as $\sum_{j \in <i>} K(x_i,u_j)$, and the normalized low-rank approximation of the affinity matrix (which is also sparse) can be calculated as $\mathbf{A}=\mathbf{Z} \mathbf{\Lambda}^{-1} \mathbf{Z}^T$ with $\mathbf{\Lambda} = diag(\mathbf{Z}^T \mathbf{1}).$ Finally, note that after using this low-rank approximation, we have $\mathbf{D}=\mathbf{I}$.

\section{Efficient Spectral Hashing}

To develop an efficient spectral hashing (ESH) algorithm, first recall that the problem in Eq. \ref{eq:2} can be written as follows:

\begin{align}\label{eq:3}
& \underset{\mathbf{B}}{\rm argmin} \quad
   -\textrm{\emph{Tr}}\{\mathbf{B}^{T}\mathbf{A}\mathbf{B}\}\\
& \text{s.t.} \quad
  \mathbf{B}^T \mathbf{B}=n \mathbf{I}_k, \quad \mathbf{B}^T \mathbf{1}_{n \times 1}=\mathbf{0}, \quad \mathbf{B}\!\in\!\{-1,1 \}^{n \times k} \nonumber
\end{align}

As pointed out, even for a single bit, this problem is extremely difficult, and continuous relaxations are generally used for simplification. However, naive relaxation generally degrades the quality of the binary codes, and for this reason, finding better continuous relaxations has been an area of study. Now, we take the first step toward efficiency. We assume that the binary codes can be obtained from the following simple model:
\begin{equation}
\mathbf{B}=sgn(\mathbf{X} \mathbf{W}),
\label{eq:4}
\end{equation}
where $\mathbf{W}$ is a $d \times k$ matrix and
$sgn(\cdot)$ denotes the element-wise sign operation. Given this model, the $n \times k$ optimization variable is replaced with the $d \times k$ matrix $\mathbf{W}$. Clearly, $d\ll n$, and thus the proposed model reduces the search space. In this case, the optimization problem in Eq.\ref{eq:3} will convert into the following problem:
\begin{equation}
\begin{aligned}
& \underset{\mathbf{W}}{\rm argmin}
& &  -\textrm{\emph{Tr}}\{sgn(\mathbf{W}^{T} \mathbf{X}^{T})\mathbf{A} sgn(\mathbf{X} \mathbf{W})\}\\
& \text{s.t.}
& & sgn(\mathbf{XW})^{T}  sgn(\mathbf{X} \mathbf{W})=n \mathbf{I}_k, \\
& & & sgn(\mathbf{XW})^{T} \mathbf{1}_{n \times 1}=\mathbf{0}.
\label{eq:5}
\end{aligned}
\end{equation}

Note that after obtaining $\mathbf{W}$, the binary matrix $\mathbf{B}$ can be calculated using Eq. \ref{eq:4}. Although the aforementioned model reduces the search space, owing to $sgn(\cdot)$, this is still a discrete optimization problem and is difficult to solve. Removing $sgn(\cdot)$ results in poor binary codes owing to the accumulated quantization error. However, if we ensure that the elements of $\mathbf{XW}$ are sufficiently close to +1 or -1, then the $\mathbf{XW} \approx sgn(\mathbf{XW})$ approximation is no longer unrealistic. To this end, we propose a novel regularization term that pushes the elements of $\mathbf{XW}$ closer to $\pm 1$ without adding any other optimization variables. This is particularly different from methods such as LGSHR, DGH, RDSH, and DSH, all of which do so by including one or more optimization variables, which leads to alternating optimization approaches. Our proposed relaxed optimization problem function has the following form:
\begin{equation}
\begin{aligned}
& \underset{\mathbf{W}}{\rm argmin}
& &  -\textrm{\emph{Tr}}\{\mathbf{W}^{T}(\mathbf{X}^{T}\mathbf{A}\mathbf{X})\mathbf{W}\} + \frac{\alpha}{2} \| \left| \mathbf{X}\mathbf{W}\right|-\mathbf{J} \| ^2_F\\
& \text{s.t.}
& &  \mathbf{W}^{T} \mathbf{X}^{T} \mathbf{X} \mathbf{W}=n \mathbf{I}_k, (\mathbf{XW})^{T} \mathbf{1}_{n \times 1}=\mathbf{0},
& & & 
\label{eq:6}
\end{aligned}
\end{equation}
where $\mathbf{J}$ is an $n \times k$ matrix with all elements equal to 1, $\left| \cdot \right|$ represents the element-wise absolute value, and $\| \cdot \|_F$ denotes the Frobenius norm. First, note that, because the data are zero-centered, the last constraint in Eq. \ref{eq:6} is already satisfied and thus we remove it from the problem statement. In addition, to further simplify the problem, we transform the orthogonality of the columns of $\mathbf{XW}$ into the orthonormality of the columns of $\mathbf{W}$. Finally, we normalize the cost function by the number of samples to obtain smoother training. Taking these changes into account and denoting $\mathbf{X}^{T}\mathbf{A}\mathbf{X}$ as $\mathbf{S}$, which is a $d \times d$ matrix, the problem in Eq. \ref{eq:6} will take the following form:
\begin{equation}
\begin{aligned}
& \underset{\mathbf{W}}{\rm argmin}
& & \mathbf{\mathcal{L}}(\mathbf{W})=\frac{-1}{n} \textrm{\emph{Tr}}\{\mathbf{W}^{T}\mathbf{S}\mathbf{W}\} + \frac{\alpha}{2n} \| \left| \mathbf{X}\mathbf{W}\right|-\mathbf{J} \| ^2_F\\
& \text{s.t.}
& &  \mathbf{W}^{T} \mathbf{W}= \mathbf{I}_k. 
\label{eq:7}
\end{aligned}
\end{equation}

To solve this constraint optimization problem, two algorithms, namely, the projected gradient and Stiefel manifold optimization, are employed.

\subsection{Projected Gradient-ESH1}
In this algorithm, during each iteration, the matrix $\mathbf{W}$ is updated using the gradient descent method as if there is no constraint, and the updated matrix is then projected to the closest matrix in the feasible set. The derivative of the first term in Eq. \ref{eq:7} is $2\mathbf{S}\mathbf{W}$. For the derivative of the second part, we first note that we have $\left| \mathbf{X}\mathbf{W}\right|=sgn(\mathbf{X}\mathbf{W})\odot \mathbf{X}\mathbf{W}$, where $\odot$ denotes the Hadamard (element-wise) product. The derivative can then be calculated as follows: 

\begin{align}
 &\frac{\partial }{\partial \mathbf{W}}   \| \left| \mathbf{X}\mathbf{W}\right|-\mathbf{J} \| ^2_{_F}\nonumber\\
 &=\frac{\partial }{\partial \mathbf{W}}Tr\left(\left(\left| \mathbf{X}\mathbf{W}\right|-\mathbf{J}\right)^T\left(\left| \mathbf{X}\mathbf{W}\right|-\mathbf{J}\right) \right) \nonumber\\
 &=2\,\mathbf{X}^T sgn\left(\mathbf{X}\mathbf{W} \right)\odot\left(\left| \mathbf{X}\mathbf{W}\right|-\mathbf{J}\right) \,\nonumber\\
 &=2\,\mathbf{X}^T\,\left( \mathbf{X}\mathbf{W}-sgn\left(\mathbf{X}\mathbf{W} \right)\right).
 \label{eq:8}
\end{align}

This derivative is valid everywhere, except for zero. For zero, we define the derivative as equal to zero. In this case, if the derivative of the cost function in Eq. \ref{eq:8} in iteration $p$ is $\mathbf{G}$, it can then be calculated as follows:

\begin{equation}
\begin{aligned}
&\mathbf{G}_p = \frac{-2}{n} \mathbf{S} \mathbf{W}_p +  \frac{\alpha}{n} \mathbf{X}^T (\mathbf{X}\mathbf{W}_p-sgn(\mathbf{X}\mathbf{W}_p)),
\label{eq:9}
\end{aligned}
\end{equation}

 In this case the learning rule for minimizing the expression can be written as follows:
\begin{equation}
\begin{aligned}
&\mathbf{W}_p =\mathbf{W}_{p-1} - \eta \mathbf{G}_{p-1},
\label{eq:10}
\end{aligned}
\end{equation}
where $\eta$ is the learning rate parameter. To complete the iteration step, we project $\mathbf{W}$ onto the feasible set. This is equivalent to finding the closest matrix $\mathbf{Q}$ to $\mathbf{W}$ such that $\mathbf{Q}^{T} \mathbf{Q}= \mathbf{I}_k$. This problem is known as the projection on the Stiefel manifold, which can be formulated as follows:

\begin{equation}
\begin{aligned}
& \textrm{Proj}(\mathbf{W}):=\underset{\mathbf{Q}}{\rm argmin} 
& & \| \mathbf{W} -\mathbf{Q} \| ^2_F\\
& \text{s.t.} \quad \mathbf{Q}^{T} \mathbf{Q}= \mathbf{I}_k.   
\label{eq:11}
\end{aligned}
\end{equation}
Fortunately, there is a closed form solution to this problem:
\begin{equation}
\mathbf{Q}=\textrm{Proj}(\mathbf{W})=\mathbf{U} \mathbf{I}_{d \times k} \mathbf{V}^T,
\label{eq:12}
\end{equation}
where $\mathbf{W}=\mathbf{U}\mathbf{\Sigma}\mathbf{V}^T$ is the SVD decomposition of $\mathbf{W}$. Interested readers can see \cite{manton2002optimization} for a detailed proof. Algorithm \ref{alg:ESH1} summarizes the proposed projected gradient method.

\begin{algorithm}[]

	\caption{The Proposed ESH1 Algorithm} 
	
	\hspace*{\algorithmicindent} \textbf{Input:} Training data $\textbf{X} \in \R ^{n\times d}$, affinity matrix $\mathbf{A} \in $  \\
	\hspace*{\algorithmicindent}$ \R ^{n\times n}$,
	number of iterations $N$, learning rate $\eta$.\\
    \hspace*{\algorithmicindent} \textbf{Output:} Binary matrix $\mathbf{B} \in \{-1,1\}^{n \times k}$. \\ 
    \hspace*{\algorithmicindent} \textbf{Initialization:} Initialize an orthogonal matrix $\mathbf{W}$: \\
    \hspace*{\algorithmicindent} $\mathbf{W}_0 \in \R ^{d \times k}$
	\begin{algorithmic}[1]
	\State $\mathbf{S} \leftarrow \mathbf{X}^T \mathbf{A}\mathbf{X}$
	\State Compute $\alpha$ according to Eq .\ref{eq:19}
	
		\For {$p=1,2,\ldots,N$}
		\State $\mathbf{W}_p \leftarrow \mathbf{W}_{p-1}-  \quad \eta \left(\frac{-2}{n} \mathbf{S} \mathbf{W}_{p-1} +  \frac{\alpha}{n} \mathbf{X}^T (\mathbf{X}\mathbf{W}_{p-1}-sgn(\mathbf{X}\mathbf{W}_{p-1})) \right)$
		\State $\textbf{U} \Sigma \textbf{V}^T \leftarrow    \textrm{SVD}(\mathbf{W}_p)$
		\State $\mathbf{Q}\leftarrow \textrm{Proj}(\mathbf{W})=\mathbf{U} \mathbf{I}_{d \times k} \mathbf{V}^T$
		\State $\mathbf{W}_p \leftarrow \mathbf{Q}$
		\EndFor
        \State $\mathbf{B} \leftarrow sgn(\mathbf{X}\mathbf{W})$
    
	\end{algorithmic} 
\label{alg:ESH1}
\end{algorithm}

\subsection{Stiefel Manifold Optimization-ESH2}

The problem in Eq. \ref{eq:7} is an optimization on the Stiefel manifold, and methods developed for the optimization on manifolds can be used to obtain a solution. In this study, we employ the method in \cite{wen2013feasible}, where an efficient algorithm has been proposed to preserve the updated $\mathbf{W}$ on the Stiefel manifold during each iteration. Briefly, to preserve the orthogonality constraint on $\mathbf{W}$ during each iteration, we define the skew-symmetric matrix $\mathbf{F}$ in iteration $p-1$ as $\mathbf{F}_{p-1} = \mathbf{G}_{p-1} \mathbf{W}_{p-1}^T - \mathbf{W}_{p-1} \mathbf{G}_{p-1}^T$, and denote the updated version of the $\mathbf{W}$ as $\mathbf{Y}(\tau)$, then using a Crank-Nicolson-like scheme we have

\begin{equation}
\mathbf{Y}(\tau)_p =\mathbf{W}_{p-1} - \frac{\tau}{2} \mathbf{F}_{p-1} \left(\mathbf{W}_{p-1}+\mathbf{Y}(\tau)_p\right),
\label{eq:13}
\end{equation}
where $\tau$ is the step size parameter. In this case, the closed-form solution for $\mathbf{Y}(\tau)$ is given as follows:

\begin{equation}
\mathbf{Y}(\tau)_p =\left(\mathbf{I} + \frac{\tau}{2} \mathbf{F}_{p-1}\right)^{-1} \left(\mathbf{I} - \frac{\tau}{2} \mathbf{F}_{p-1}\right)\mathbf{W}_{p-1}.
\label{eq:14}
\end{equation}

Following \cite{wen2013feasible}, we employ the Barzilai-Borwein (BB) method to reduce the total number of iterations:

\begin{equation}
\tau_p =\frac{|\textrm{\emph{Tr}}\left((\mathbf{M}_{p})^T(\mathbf{Y}_{p}) \right)|}{\textrm{\emph{Tr}}\left((\mathbf{Y}_{p})^T(\mathbf{Y}_{p}) \right)},
\label{eq:15}
\end{equation}
where $\mathbf{M}_{p}=\mathbf{W}_{p}-\mathbf{W}_{p-1}$, and $\mathbf{Y}_{p}=\nabla \mathbf{\mathcal{L}}(\mathbf{W}_p)-\nabla \mathbf{\mathcal{L}}(\mathbf{W}_{p-1})$, in which $\nabla \mathbf{\mathcal{L}}(\mathbf{W})=\mathbf{G}-\mathbf{W}\mathbf{G}^T\mathbf{W}$ is the gradient of the loss function in the tangent planes.
Algorithm \ref{alg:ESH2} summarizes the proposed manifold optimization method.

\begin{algorithm}[]

	\caption{The Proposed ESH2 Algorithm} 
	
	\hspace*{\algorithmicindent} \textbf{Input:} Training data $\textbf{X} \in \R ^{n\times d}$, affinity matrix $\mathbf{A} \in $  \\
	\hspace*{\algorithmicindent}$ \R ^{n\times n}$,
	number of iterations $N$.\\
    \hspace*{\algorithmicindent} \textbf{Output:} Binary matrix $\mathbf{B} \in \{-1,1\}^{n \times k}$. \\ 
    \hspace*{\algorithmicindent} \textbf{Initialization:} Initialize an orthogonal matrix $\mathbf{W}$: \\
    \hspace*{\algorithmicindent} $\mathbf{W}_0 \in \R ^{d \times k}$, step size $\tau$.
	\begin{algorithmic}[1]
	\State $\mathbf{S} \leftarrow \mathbf{X}^T \mathbf{A}\mathbf{X}$
	\State Compute $\alpha$ according to Eq .\ref{eq:19}
	
		\For {$p=1,2,\ldots,N$}
		\State $\mathbf{G}_{p-1} \leftarrow  \frac{-2}{n} \mathbf{S} \mathbf{W}_{p-1} +  \frac{\alpha}{n} \mathbf{X}^T (\mathbf{X}\mathbf{W}_{p-1}-sgn(\mathbf{X}\mathbf{W}_{p-1}))$
		\State $\mathbf{F}_{p-1} \leftarrow  \mathbf{G}_{p-1} \mathbf{W}_{p-1}^T - \mathbf{W}_{p-1} \mathbf{G}_{p-1}^T$
		\State $\mathbf{Y}(\tau)_p  \leftarrow  \left(\mathbf{I} + \frac{\tau}{2} \mathbf{F}_{p-1}\right)^{-1} \left(\mathbf{I} - \frac{\tau}{2} \mathbf{F}_{p-1}\right)\mathbf{W}_{p-1}$
		\State $\mathbf{W}_p\leftarrow \mathbf{Y}(\tau)_p$
		\State $\tau \leftarrow \frac{|\textrm{\emph{Tr}}\left((\mathbf{M}_{p})^T(\mathbf{Y}_{p}) \right)|}{\textrm{\emph{Tr}}\left((\mathbf{Y}_{p})^T(\mathbf{Y}_{p}) \right)}$
		\EndFor
        \State $\mathbf{B} \leftarrow sgn(\mathbf{X}\mathbf{W})$
    
	\end{algorithmic} 
\label{alg:ESH2}
\end{algorithm}

\subsection{Out of Sample: Hashing New Data}
Thus far, we have derived a binary representation for the training data. To obtain a binary representation for a new data point $\mathbf{x^*}$, the same approach as in many spectral hashing methods can be used \cite{liu2014discrete, li2017large}. More precisely, let $\mathbf{b(x^*)}$ be the binarized version of $\mathbf{x^*}$. Then, using a similar approach as applied for training, we can write the following:
\begin{equation}
\begin{aligned}
& \underset{\mathbf{b(x^*)}}{\rm argmin} 
& & \sum_{i}^n A(\mathbf{x}_i,\mathbf{x^*}) \lVert \mathbf{b}_i -  \mathbf{b(x^*)} \rVert_2 ^2, \\
& \text{s.t.} 
& &  \mathbf{b(x^*)} \in \{-1,1 \}^k
\label{eq:16}
\end{aligned},
\end{equation} 
where $A(\mathbf{x}_i,\mathbf{x^*})=\mathbf{z}_i \Lambda^{-1} \mathbf{z(x^*)}$, and $\mathbf{z}_i$ is the $i$-th row of matrix $\mathbf{Z}$. By expanding $\lVert \mathbf{b}_i -  \mathbf{b(x^*)} \rVert_2 ^2$, this problem can be written as
\begin{equation}
\begin{aligned}
& \underset{\mathbf{b(x^*)}}{\rm argmax} 
& & \langle \mathbf{b(x^*)}, \mathbf{B}^T \mathbf{Z} \Lambda^{-1}\mathbf{z(x^*)} \rangle \\
& \text{s.t.} 
& &  \mathbf{b(x^*)} \in \{-1,1 \}^k.
\label{eq:17}
\end{aligned}
\end{equation} 
In this case, the solution is 

\begin{equation}
\mathbf{b(x^*)}=sgn(\mathbf{B}^T \mathbf{Z} \Lambda^{-1}\mathbf{z(x^*)}).
\label{eq:18}
\end{equation} 

\section{Experiments}
\subsection{Datasets and Evaluation Protocol}

The proposed ESH1 and ESH2 algorithms were validated on four benchmark datasets, namely, CIFAR-10 \cite{krizhevsky2009learning}, NUS-WIDE \cite{10.1145/1646396.1646452}, LabelMe-12-50K \cite{uetz2009large}, and a set of medical images NCT-CRC-HE-100K \cite{macenko2009method}. 

The \textbf{CIFAR-10} contains 60,000 color images with a pixel resolution of $32 \times 32$ from 10 classes. The \textbf{NUS-WIDE} dataset covers 269,000 images collected from Flickr. This dataset is multi-label and contains 81 ground-truth concepts. The \textbf{LabelMe-12-50K} dataset, which is highly imbalanced, has 12 classes and consists of 50,000 images with a pixel resolution of $256\times 256$. The images have multiple label values of between zero and 1. Here, following the common setting \cite{He_2019_CVPR}, the most probable class is considered as the image label. Finally, the \textbf{NCT-CRC-HE-100K} dataset is a 9-class dataset that includes 100,000 non-overlapping image patches of colorectal cancer and normal tissue with a pixel resolution of $224\times 224$. 

Standard measures were employed to validate the performance of the ESH. These criteria involve the \emph{mean average precision } (mAP), \emph{precision at $N$} samples (e.g., precision@1000), and \emph{precision at radius 2} (precision@r=2), which calculates the precision for retrieving images with a Hamming distance equal to or less than 2 from the query image. If no image is found at radius 2, the precision is considered to be zero for that query. Because LabelMe-12-50K is an imbalanced dataset, following settings from previous studies, the \emph{macro average} mAP was reported to steer clear of the bias.

\subsection{Implementation note}
To avoid tuning the parameters, we set the learning rate parameter for all datasets and experiments to a fixed value ($\eta=0.01$). Furthermore, we propose a novel method for automatically determining the regularization parameter $\alpha$ for each dataset. To this end, we call the first and second terms in Eq. \ref{eq:7} $\mathbf{T}_1$ and $\mathbf{T}_2$, respectively. If we denote the initialization of $\mathbf{W}$ as $\mathbf{W}^{0}$, then we propose setting $\alpha$ such that the importance of the first and second parts of the cost function is the same in the first iteration:
\begin{equation}
\left|\mathbf{T}_1(\mathbf{W}^{0})\right|=\frac{\alpha}{2} \left| \mathbf{T}_2(\mathbf{W}^{0}) \right| \Rightarrow \alpha= \left| \frac{2\mathbf{T}_1(\mathbf{W}^{0})}{\mathbf{T}_2(\mathbf{W}^{0})} \right|.
\label{eq:19}
\end{equation}

For a low-rank affinity construction, following the common settings used in studies on graph hashing, we set $m = 300$ and $s = 3$ \cite{hu2018discrete, liu2014discrete}. The ESH code is provided in the Supplementary Material section.


\begin{table}[hbt]
\caption{Comparison of retrieval performance, for 16, 32 and 64 bits based on macro mAP (average over classes) for LabelMe-12-50k dataset represented by 4096-D VGG-FC7 descriptor. The best performance values are  highlighted in boldface.}
\setlength{\tabcolsep}{6pt}
\centering
\begin{tabular}{|l|cccc|}
\hline
                                     & \multicolumn{4}{c|}{macro mAP \%}                                                                      \\ \hline                                                                                   
Method                        & \multicolumn{1}{c|}{16 Bit}       & \multicolumn{1}{c|}{32 Bit}       & \multicolumn{1}{c|}{64 Bit}  &\multicolumn{1}{c|}{128 Bit}         \\ \hline\hline                             
 SH                                    & 12.60                & 12.59           & 12.24          & -                                                    \\ \hline
 
 SpH                          & 13.59                  &15.10         &17.03   & -                                                                                                    \\ \hline
KMH                                              &13.36            &15.47                 &16.58        & -                                                                     \\ \hline

BA                                   & 16.96         &18.42       &20.80             & -                           \\ \hline

ITQ                       & 17.61              & 18.65                &20.10      &21.49                                                   \\ \hline
DGH     &21.45    	&22.74	   &25.41   	&26.77
\\ \hline  

LGHSR  &21.10	&23.49	&23.98	&22.85
\\\hline 

KNNH                    &20.13           &23.34                      &26.06                 & 27.62                                              
                                                           \\ \hline
DSH    &\textbf{24.70}	 &23.78	   &24.35	 &22.39    
 \\ \hline

SCQ    &22.89	 &24.95	   &26.50	 &26.35    
 \\ \hline
 
ESH1             &22.95                            & 25.67                                                 & 27.59        & 28.94                        \\ \hline 

ESH2             &22.87   &\textbf{26.85}   &\textbf{28.20}   &\textbf{29.61}

                               \\ \hline

\end{tabular}

\label{table:table1}
\end{table}
\begin{table*}[hbt]
\centering
\setlength{\tabcolsep}{4pt}
\caption{Comparison of retrieval performance, based on mAP, precision@1000, and precision@r=2 on CIFAR-10 represented by 4096-D VGG-FC7 features. The best performance values are  highlighted in boldface.}
\begin{tabular}{|l|cccc|cccc|ccc|}
\hline
                  & \multicolumn{4}{c|}{mAP \%}    & \multicolumn{4}{c|}{precision \% @1000} & \multicolumn{3}{c|}{precision@r=2}                                                                                                                                                                           \\ \hline
Method

& \multicolumn{1}{c|}{16 Bits}       & \multicolumn{1}{c|}{32 Bits}       & \multicolumn{1}{c|}{64 Bits}  &\multicolumn{1}{c|}{128 Bits}      & \multicolumn{1}{c|}{16 Bits}  & \multicolumn{1}{c|}{32 Bits}       & \multicolumn{1}{c|}{64 Bits}    &\multicolumn{1}{c|}{128 Bits} & \multicolumn{1}{c|}{16 Bits}  & \multicolumn{1}{c|}{32 Bits}  &\multicolumn{1}{c|}{64 Bits}    \\ \hline\hline                             
 SH                          & 18.31                                  &16.54                             &15.78                                & -     & -    & -    & -   & -   & -       & -       & -                                      \\ \hline
 SpH                        & 18.82                            &20.93                                   & 23.40                                & -    & -    & -   & -   & -   & -  & -     & -                                                 \\ \hline
KMH                           & 18.68                                   &20.82                       &22.87                                   & -    & -   & -  
& -   & -     & -   & -     & -                                        \\ \hline
BA                        & 25.38                         &26.16                             & 27.99                                  & -      & -    & -    & -    & -  & - & -    & -                                                \\ \hline

ITQ             &26.23                            & 26.73                              & 27.90                              & 29.22       & 36.71	      & 38.79   &40.75	   &42.53  & \textbf{39.33}	    & 27.71  & 00.06                                 \\ \hline

DGH             & 27.73         & 27.44                            &28.01       & 29.47       & 40.36      & 39.46  &39.70	  &40.69  &38.73	    & 42.25        & 44.04	                         \\ \hline

LGSHR &27.83  &25.87	&22.12	&19.66   &39.98	  &42.84  	&41.82	&40.45   &38.45	  &46.11	&32.75	  

\\ \hline

KNNH         & 29.25                          &  30.55                               &32.60                              &33.68         &38.13	 &40.51	   &43.32    &	44.57      & 37.66	       &24.95	       & 03.43	                      \\ \hline                                                       
DSH             &27.72	              & 25.36                             &22.12	                             & 19.55      & 40.36	    & 42.87    &	41.80  &	40.57  & 38.73	        & 45.69   & 	26.15	                               \\ \hline

SCQ             &27.52	              & 27.42                             &30.34	                             & 32.25      & 34.24	    & 36.51    &	40.29  &	43.15  & 30.58	        & 38.24   & 	25.14	                               \\ \hline

ESH1            &\textbf{32.11}                            & 33.08                              & 34.47                           & 34.92         & 38.67       & 42.15  &44.75      &45.70    & 35.03    &43.59    & \textbf{44.26}                   \\ \hline

ESH2            &31.59                           & \textbf{33.44}                              & \textbf{34.72}                           & \textbf{35.29}         & \textbf{41.32}       &\textbf{43.47}   &\textbf{45.16}      &\textbf{45.95}    & 38.73    & \textbf{46.69}    & 43.58              \\ \hline

\end{tabular}

\label{table:table2}
\end{table*}

\subsection{Results on LabelMe-12-50K}
We report the macro mAP for this imbalanced dataset. A VGG network was employed for the feature extraction. For the test \& train split, similar to the common setting, we sample 10\% of each class as the test data and 90\% as the training set. Table \ref{table:table1} presents the performance of ESH1 and ESH2 compared with the other methods. It is obvious that, except for the 16-bit setting where DSH obtains a better performance, for the 32, 64, and 128 bits, ESH1 and ESH2 attain a better macro mAP compared to other algorithms. In Table \ref{table:table1}, SpH indicates spherical hashing \cite{heo2012spherical}, and KMH stands for k-means hashing \cite{he2013k}.

\begin{table*}[hbt]
\centering
\setlength{\tabcolsep}{4pt}
\caption{Comparison of retrieval performance based on mAP, precision@5000, and precision@r=2 on NUS-WIDE dataset represented by VGG-F deep features. The best performance values are highlighted in boldface.}
\begin{tabular}{|l|cccc|cccc|ccc|}
\hline
                  & \multicolumn{4}{c|}{mAP \%}    & \multicolumn{4}{c|}{precision \% @5000}    
                  & \multicolumn{3}{c|}{precision@r=2} 
                  \\ \hline
Method                    & \multicolumn{1}{c|}{16 Bits}       & \multicolumn{1}{c|}{32 Bits}       & \multicolumn{1}{c|}{64 Bits}  & \multicolumn{1}{c|}{128 Bits}     & \multicolumn{1}{c|}{16 Bits}  & \multicolumn{1}{c|}{32 Bits}       & \multicolumn{1}{c|}{64 Bits}     & \multicolumn{1}{c|}{128 Bits}   & \multicolumn{1}{c|}{16 Bits}  & \multicolumn{1}{c|}{32 Bits}       & \multicolumn{1}{c|}{64 Bits}       \\ \hline \hline  
LSH                       & 40.45                                   & 48.04                               & 46.75         &-                        &47.95     &55.29         &58.95    &-   &-  &-  &-                                        
                                   \\ \hline
 SH                        &44.74                                    &42.60                                &42.36           &-                      & 59.15    & 54.98    & 54.18      &-                &-  &-  &-                                                \\ \hline

AGH                         & 49.80        &47.14                             & 44.72     &-                             & 70.43    & 70.29   & 69.29  &-  &-  &-   &-  
                                              \\ \hline
                                              
SGH                     &48.86                                   &49.10                                & 51.33        &-                & 64.92
& 66.04   & 69.01      &-    &-    &-    &-                                               \\ \hline
ITQ         &52.09	&53.12	&54.04	&54.85	&64.34	&66.62	&68.48	&69.54	&67.31	&51.91	&05.21	                                        \\ \hline

DGH                      &51.35	  &52.21	&53.34	  &55.96	&66.33	&65.35	&66.02	&67.20	&65.14	&68.41	&\textbf{70.19}	                                                  \\ \hline

LGHSR              &50.06	&47.72	&45.71	&44.11	&\textbf{68.42}	&68.09	&67.92	&66.95	&\textbf{68.87}	&71.98	&45.27	                                                \\ \hline

KNNH    &55.12	&57.03	&\textbf{58.61}	&\textbf{59.45}	&66.77	&69.83	&71.07	&72.13	&68.61	&49.78	&09.58	
           \\ \hline 
           
DSH        &49.87	&47.07	&45.67	&44.03	&68.05	&68.17	&67.83	&67.15	&68.67	&71.86	&39.93     
  \\ \hline 

SCQ            &56.34  	 &56.21	   &56.10	 &53.54	   &68.45	  &70.64	 &70.81	   &69.00	 &68.74   &70.92	 &08.98	                   \\ \hline
ESH1            &56.32  	 &56.89	   &57.47	 &57.28	   &67.58	  &69.47	 &71.46	   &72.31	 &64.64	   &\textbf{72.30}	 &60.14	                   \\ \hline

ESH2            &\textbf{56.54}   &\textbf{57.16}    &57.71  &57.53   &68.20   &\textbf{70.80}      &\textbf{72.14}  &\textbf{72.45}  &64.34  &72.06  & 62.08 

              \\ \hline
\end{tabular}
\label{table:table3}
\end{table*}

\subsection{Results on CIFAR-10}

For CIFAR-10, similar to \cite{He_2019_CVPR}, each image is represented by a deep 4096-D feature extracted from the VGG network \cite{simonyan2014very}. For the test \& train split, $10\%$ of each class is sampled as the query set, and the remaining instances are sampled as the training set. Table \ref{table:table2} shows the results for CIFAR-10 based on mAP, precision@1000, and precision@r=2. As can be seen, ESH1 and ESH2 outperform the state-of-the-art approaches, namely, LGHSR, KNNH, and DSH, based on mAP and precision@1000 for all 16-, 32-, 64-, and 128-bit settings. For precision@r=2, ESH1 and ESH2 achieved the best performance for 64 and 16 bits, respectively, and competitive results for 32 bits. The first row in Fig. \ref{fig:1} compares the performance of the ESH algorithms with recent competitive algorithms, LGHSR, DSH, and KNNH based on precision-recall graphs. Clearly, the ESH algorithms achieved a better performance than the recent methods.

\subsection{Results on NUS-WIDE}
\begin{figure*}[h]
  \centering

  \medskip

  \begin{subfigure}[t]{.3\linewidth}
    \centering\includegraphics[width=5cm]{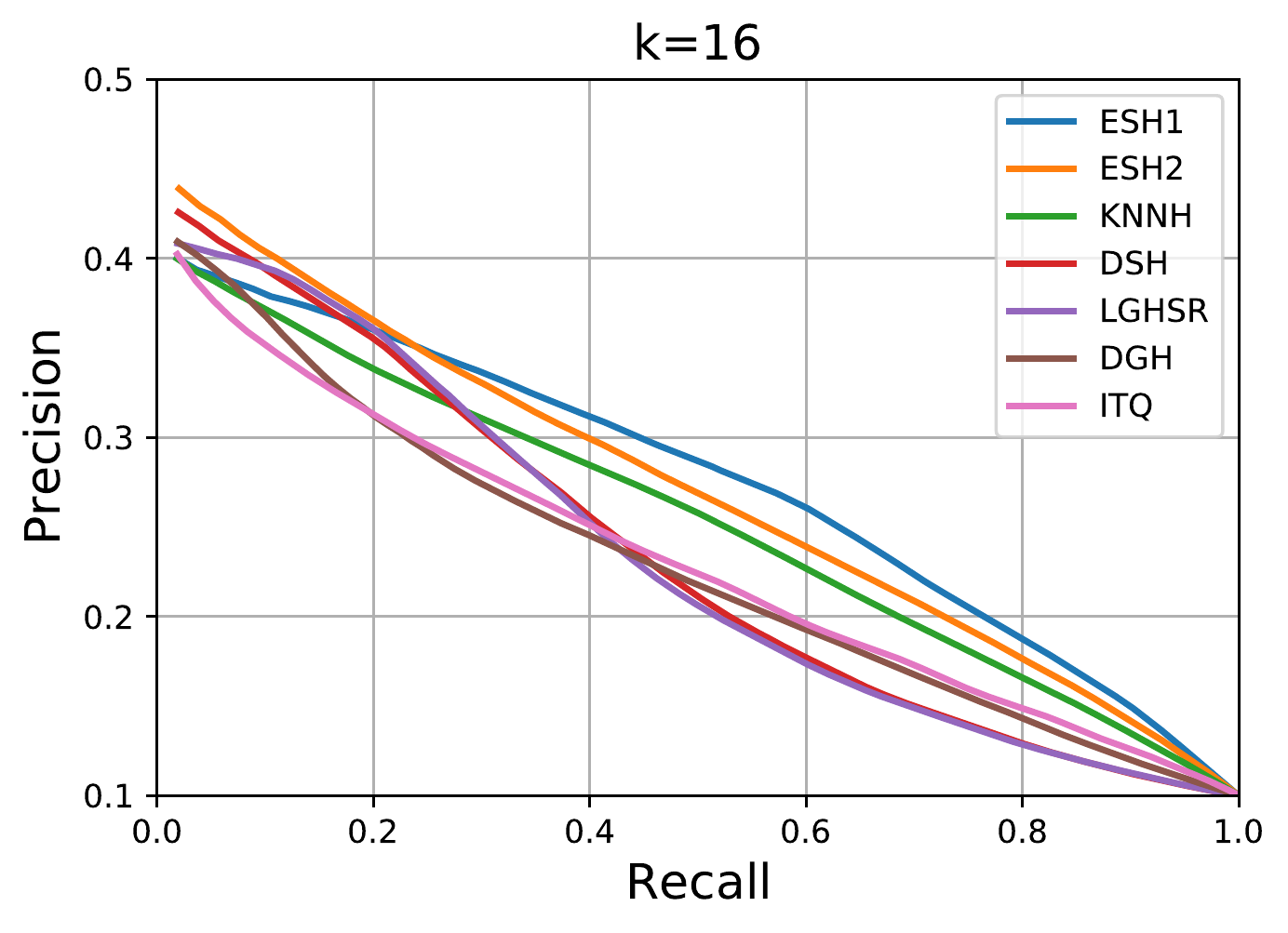}
    \caption{}
  \end{subfigure} \hspace{5mm}
  \begin{subfigure}[t]{.3\linewidth}
    \centering\includegraphics[width=5cm]{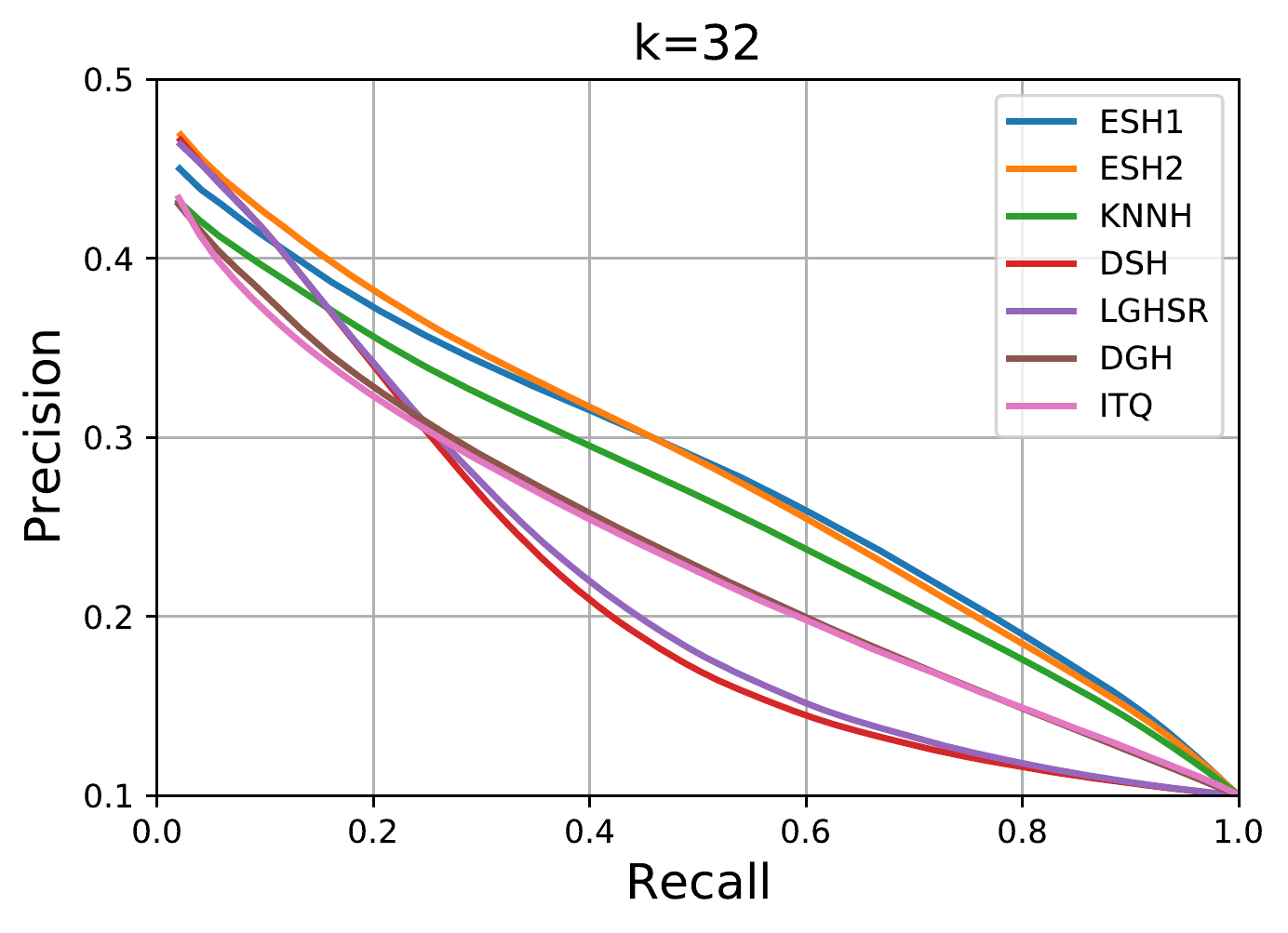}
    \caption{}
  \end{subfigure} \hspace{5mm}
  \begin{subfigure}[t]{.3\linewidth}
    \centering\includegraphics[width=5cm]{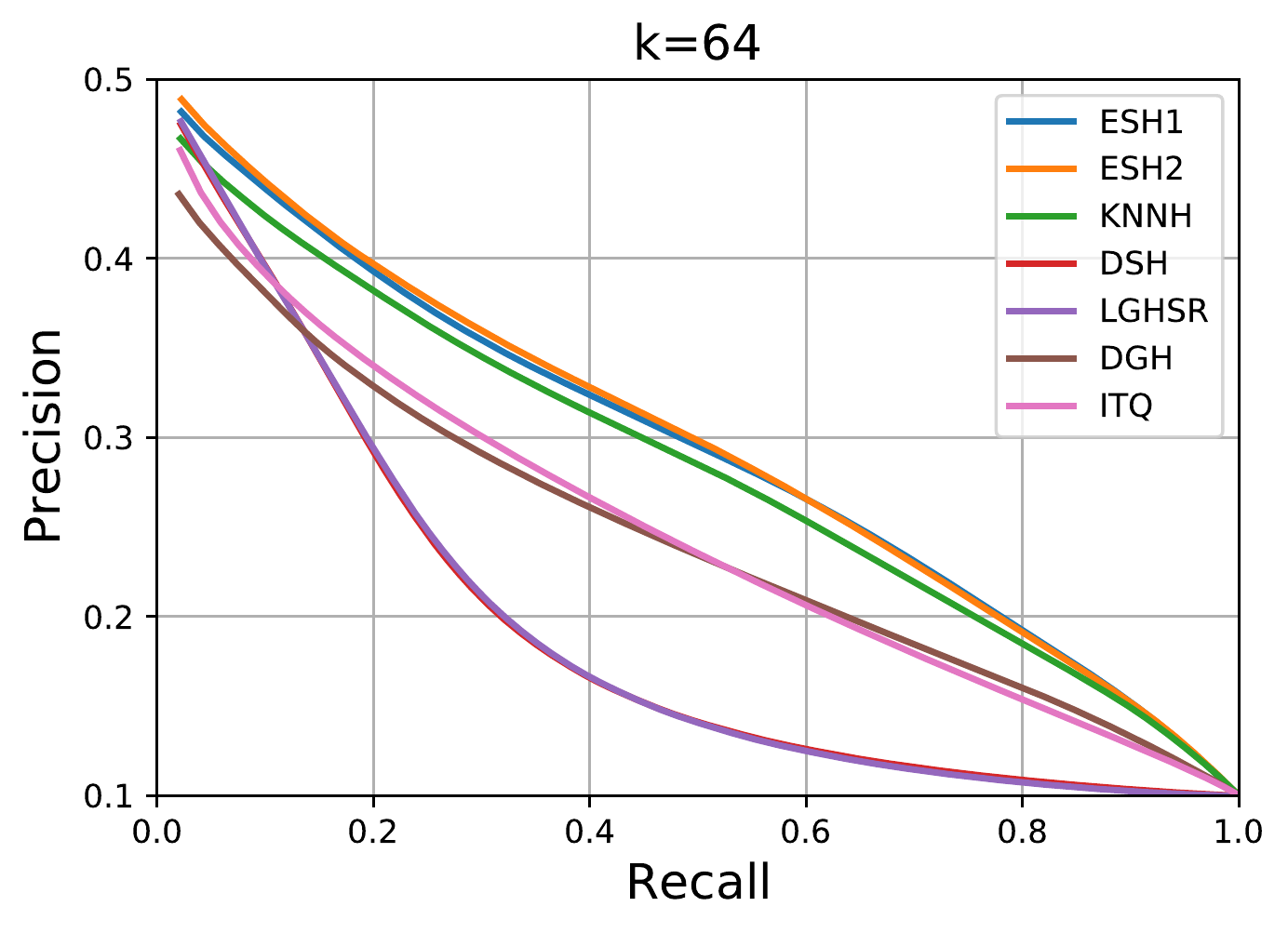}
    \caption{}
    \end{subfigure}
    \begin{subfigure}[t]{.3\linewidth}
    \centering\includegraphics[width=5cm]{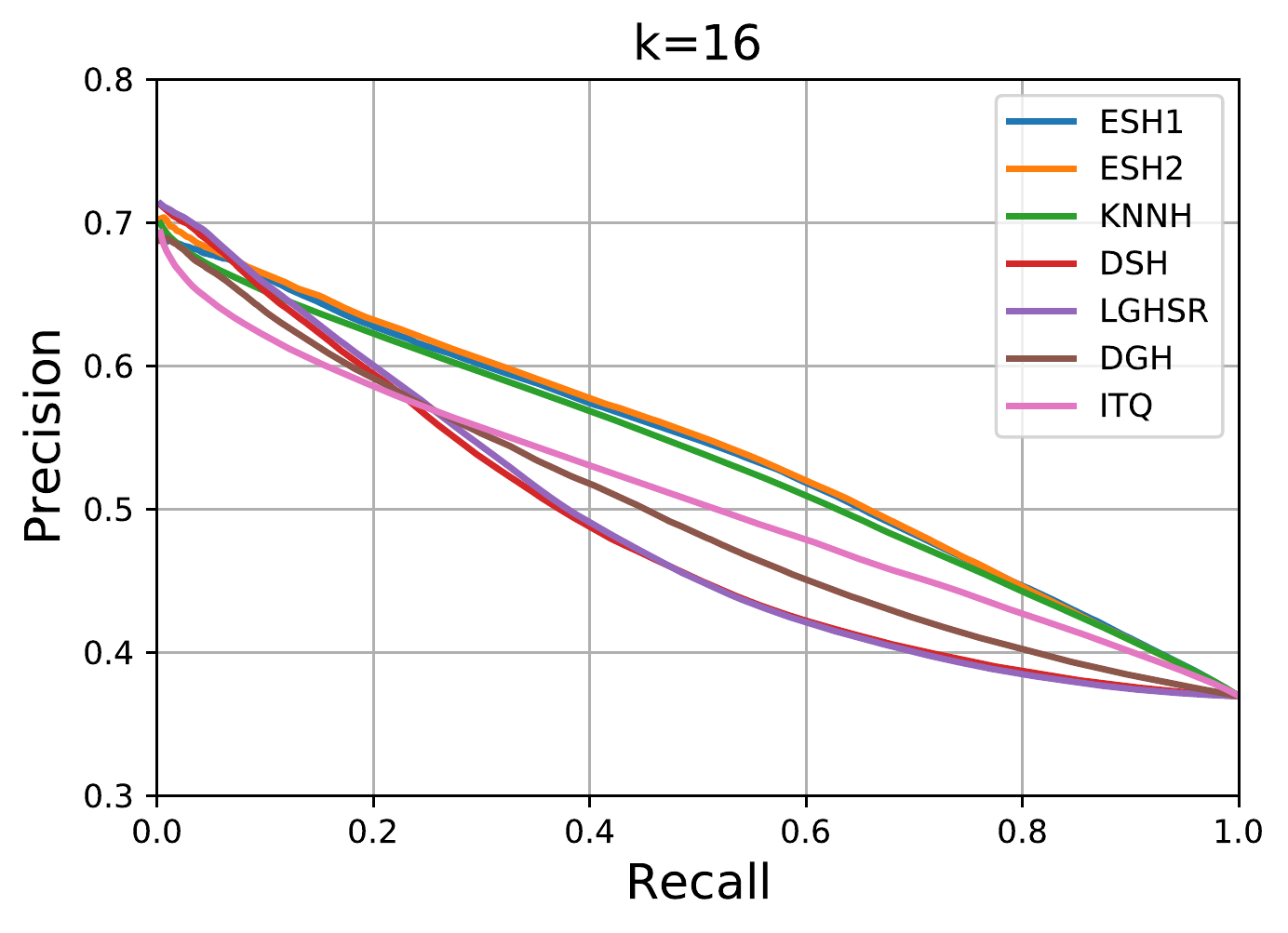}
    \caption{}
    \end{subfigure} \hspace{5mm}
     \begin{subfigure}[t]{.3\linewidth}
    \centering\includegraphics[width=5cm]{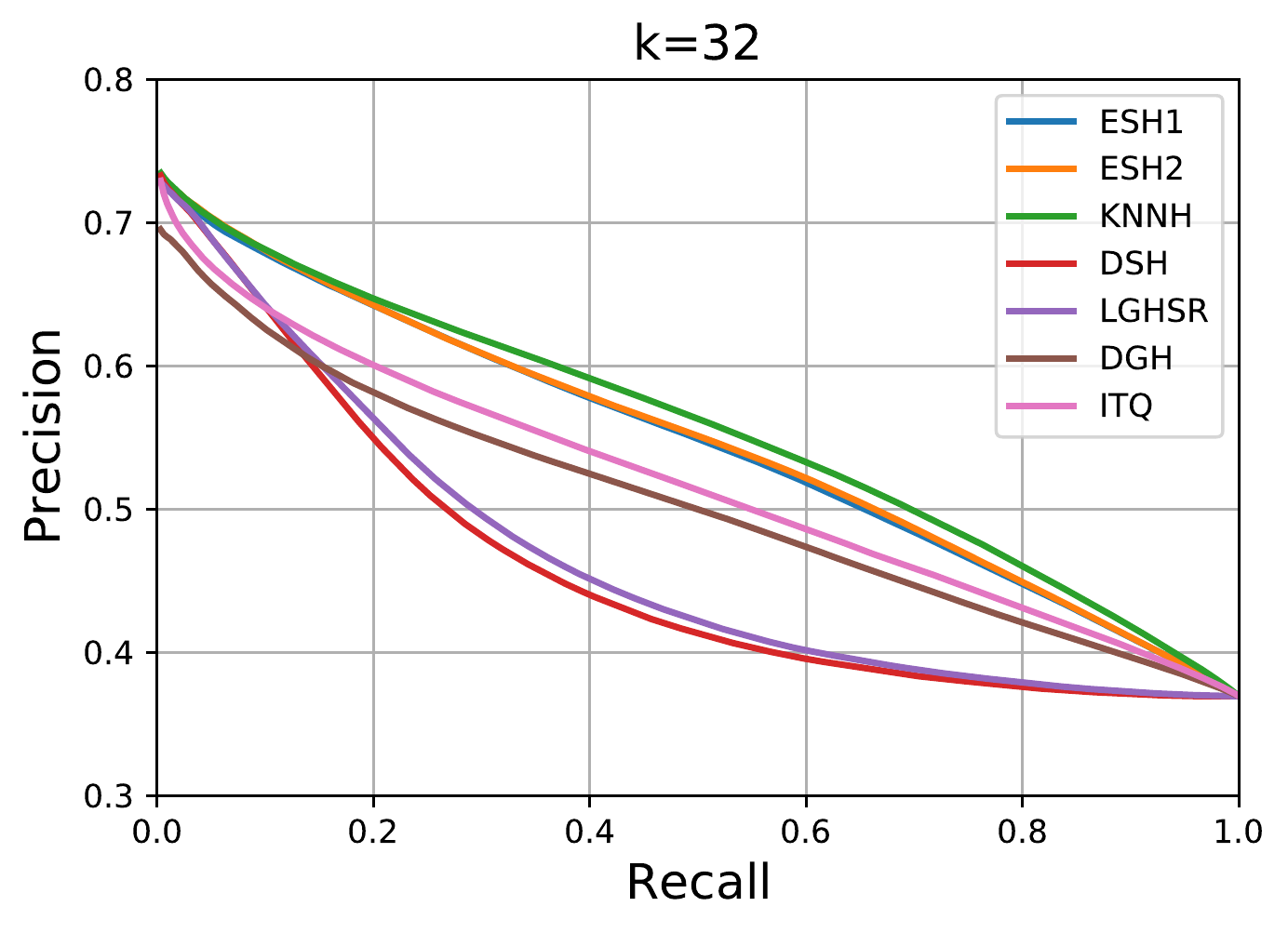}
    \caption{}
     \end{subfigure} \hspace{5mm}
     \begin{subfigure}[t]{.3\linewidth}
    \centering\includegraphics[width=5cm]{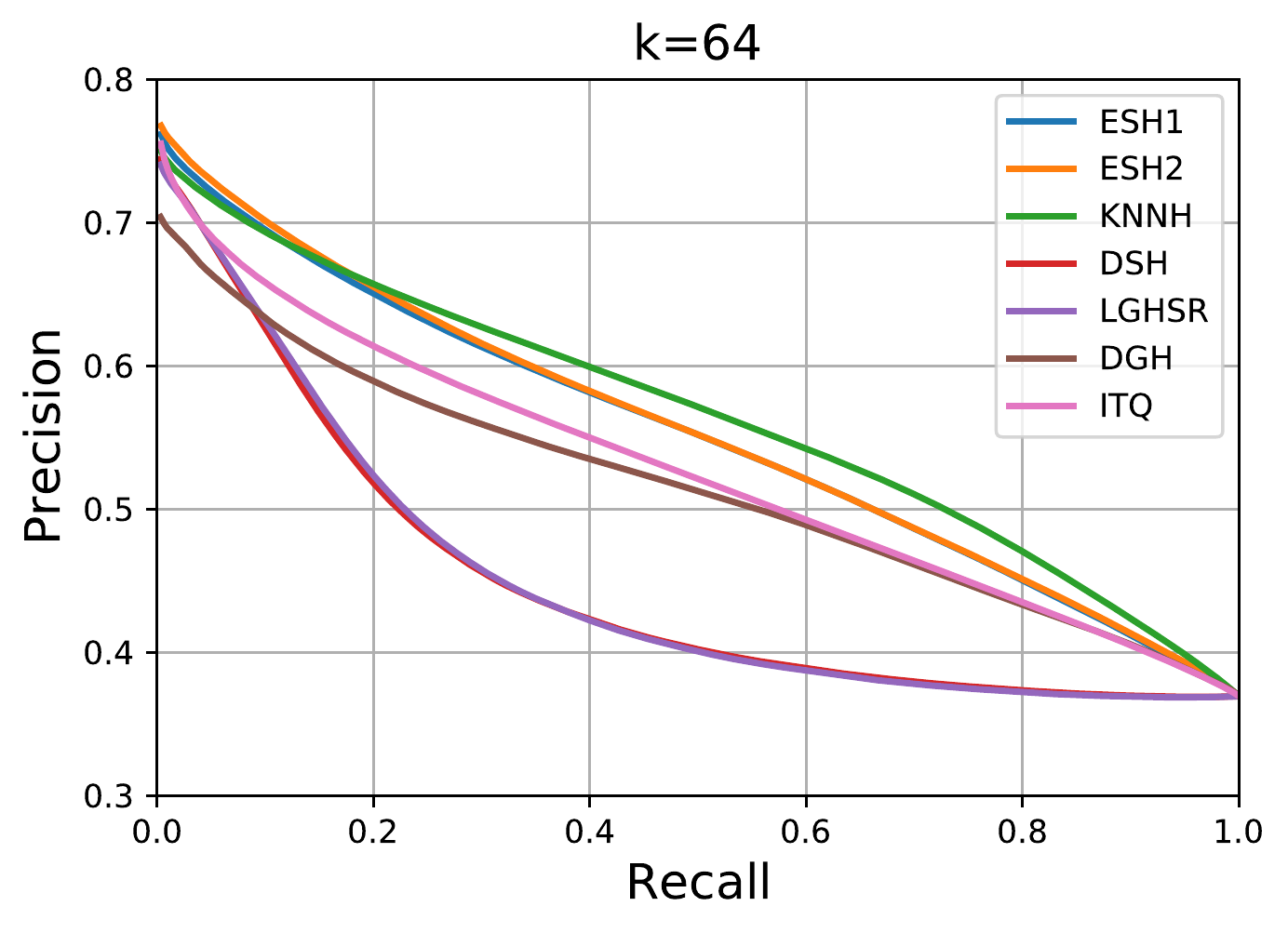}
    \caption{}
    \end{subfigure}
    \begin{subfigure}[t]{.3\linewidth}
    \centering\includegraphics[width=5cm]{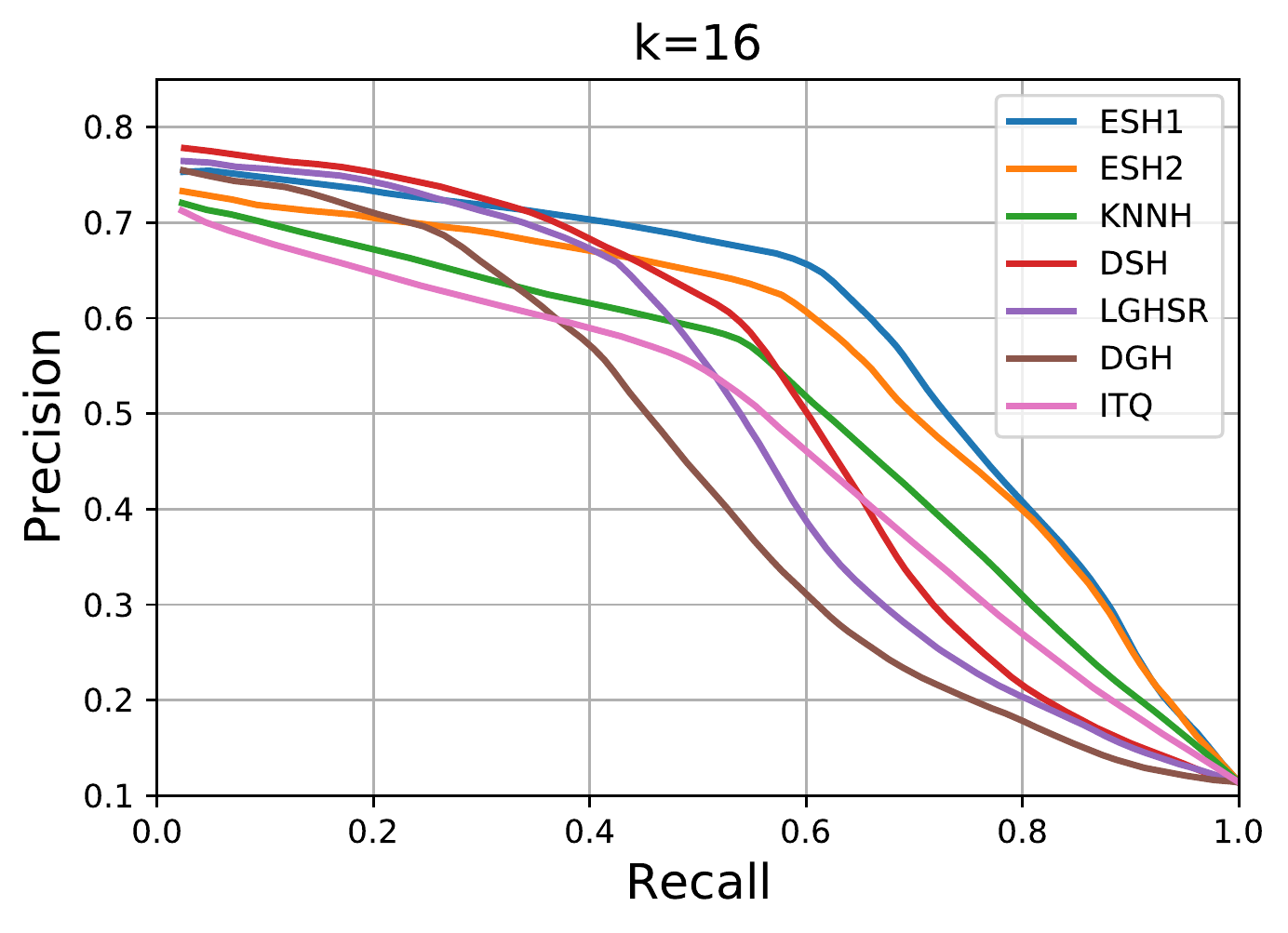}
    \caption{}
  \end{subfigure} \hspace{5mm}
  \begin{subfigure}[t]{.3\linewidth}
    \centering\includegraphics[width=5cm]{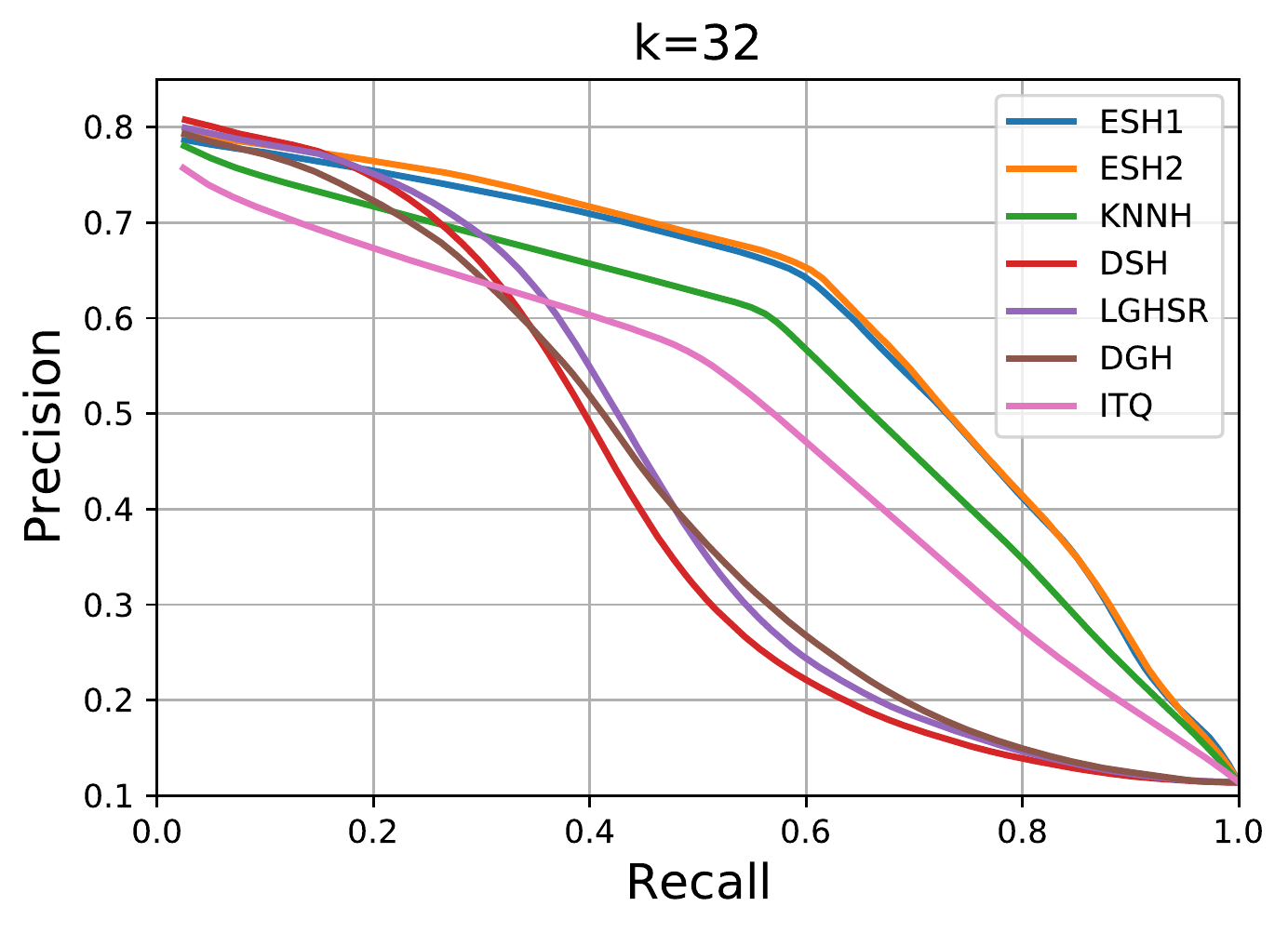}
    \caption{}
  \end{subfigure} \hspace{5mm}
  \begin{subfigure}[t]{.3\linewidth}
    \centering\includegraphics[width=5cm]{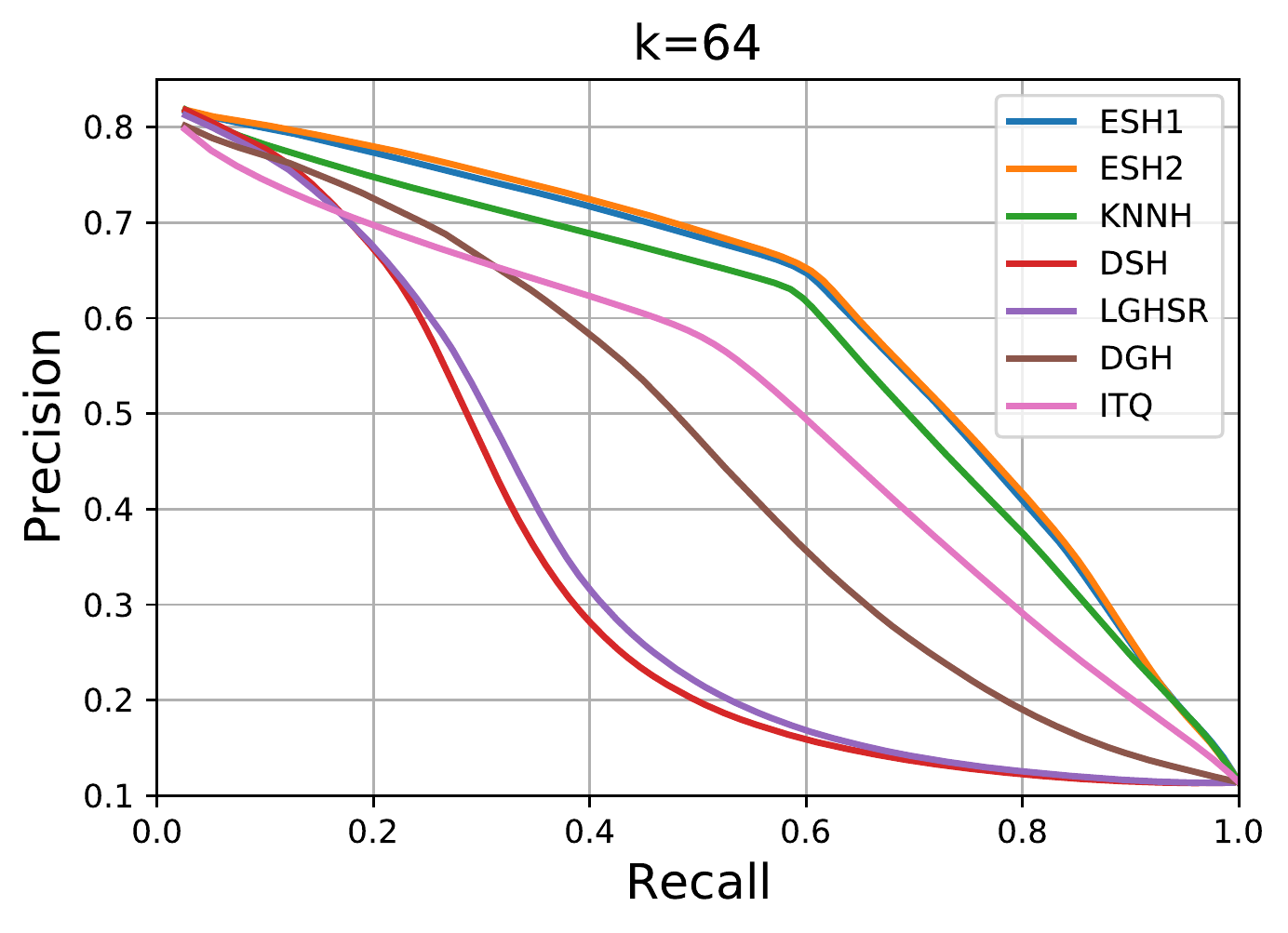}
    \caption{}
    \end{subfigure}
  \caption{Precision-recall graphs for CIFAR-10 (first row), NUS-WIDE (second row), and NCT-CRC-HE-100K (third row) for different numbers of bits ($k = 16, 32, 64$).}%
  \label{fig:1}%
\end{figure*}

For this multi-label dataset, similar to the common setting \cite{liu2011hashing}, VGG features were used for image representation. Images with labels among the 21 most frequent labels (195,834 images) were selected. We randomly sampled 100 images from each class to construct the test set, and the remaining images were used to train the hash function and populate the hash table. For the mAP and precision calculation, two images are considered neighbors if they share at least one common label. Table \ref{table:table3} presents the validation of the ESH1 and ESH2 algorithms on this dataset. Here, LSH indicates locality sensitivity hashing \cite{gionis1999similarity}, PCAH is semi-supervised hashing \cite{wang2012semi}, and SGH represents salable graph hashing \cite{jiang2015scalable}. For mAP, ESH1 and ESH2 outperform DSH, which is the most recent method based on a spectral hashing formulation and has a higher complexity than ESH. For the NUS-WIDE dataset, in some cases, ESH algorithms do not outperform  the state-of-the-art method KNNH. Note that with the KNNH method, there exists an $O(n^2 d)$ complexity for distance computing and an $O(n^2\log_2 n)$ complexity for sorting, whereas the ESH methods are by far more efficient with a complexity of $O(2ndkN+ndm)$. Furthermore, for precision@r=2, compared with DGH, DSH, and LGHSR, the results of the ESH methods do not always show the best performance. In this regard, first, we should note that all DGH, DSH, and LGHSR have been formulated based on two decision variables (one discrete and one continuous), whereas the ESH is a function of one decision variable. This means that the runtime of the ESH methods are significantly lower (see Fig. \ref{fig:2} ) compared with these methods. Considering this lower complexity, providing a competitive performance compared with the state-of-the-art approaches imply that the ESH algorithms propose a tradeoff between lower complexity and simultaneously achieving a good performance. Second, for precision@r=2, it is true that ESH values are not always the state-of-the-art, but we should note that, in precision@r=2, r = 2 is an empirical setting for reducing the number of retrieved images, which leads to the different performances achieved by the different methods. Finally, it is clear from the precision-recall graphs in the second row of Fig. \ref{fig:1} that the ESH algorithms are competitive compared with the state-of-the-art approaches.

\begin{table*}[h]
\centering
\setlength{\tabcolsep}{4pt}
\caption{Comparison of retrieval performance based on mAP, precision@1000, and precision@r=2 on NCT-CRC-HE-100K dataset represented by  EfficientNet features. The best performance values are highlighted in boldface.}
\begin{tabular}{|l|cccc|cccc|ccc|}
\hline
                  & \multicolumn{4}{c|}{mAP \%}    & \multicolumn{4}{c|}{precision \% @1000} & \multicolumn{3}{c|}{precision@r=2}                                                                                                                                                                           \\ \hline
Method                    & \multicolumn{1}{c|}{16 Bits}       & \multicolumn{1}{c|}{32 Bits}       & \multicolumn{1}{c|}{64 Bits}  & \multicolumn{1}{c|}{128 Bits}    & \multicolumn{1}{c|}{16 Bits}  & \multicolumn{1}{c|}{32 Bits}       & \multicolumn{1}{c|}{64 Bits}     & \multicolumn{1}{c|}{128 Bits} & \multicolumn{1}{c|}{128 Bits}  & \multicolumn{1}{c|}{32 Bits}  &\multicolumn{1}{c|}{64 Bits}       \\ \hline\hline  

ITQ            & 54.72                             & 55.78                            & 57.50   &58.39                           & 68.41      & 71.64      & 74.63    &76.14    &67.90    &72.54    &35.66                                    \\ \hline

DGH 
  &49.85	 &47.24	   &51.61	   &59.61	   &74.08	 &77.14	   &77.05	 &77.58	    &68.75	 &79.02	   &79.15	 

\\ \hline

LGHSR   &55.28	&47.74	&38.65	&32.37	&75.66	&78.21	&77.13	&74.67	&73.80	  &79.82	  &67.76
\\ \hline

KNNH             &58.02	  &61.47	  &64.28	&  65.79	&70.27	  &74.91	&78.16	  &80.30	&68.37  	&69.56	&42.73	                                        \\ \hline                                        

DSH   &58.75	&45.63	&37.27	&32.52	&\textbf{76.70}	&78.71	&77.73	&75.30	&\textbf{75.15}	&79.56	&61.49	
\\ \hline

SCQ   &64.90	 &67.25	&67.75	&66.57	&76.03	&\textbf{80.07}	&80.01	&80.02	&69.64	&\textbf{80.51}	&72.65	

\\ \hline
ESH1            &\textbf{66.32}	  &66.77	&67.14	&66.26	&74.82	&77.36	&79.84	&80.32	&67.14	&76.92	&80.02	              \\ \hline

ESH2
   &63.54 	&\textbf{67.30}   &\textbf{67.75}   &\textbf{67.04}   &71.85   &78.12   &\textbf{80.16}  &\textbf{80.46}   &63.04  &77.17  &\textbf{80.41}   
\\ \hline
\end{tabular}

\label{table:table4}
\end{table*}

\subsection{Results on NCT-CRC-HE-100K}
For the NCT-CRC-HE-100K dataset, EfficientNet \cite{tan2019efficientnet} pre-trained on ImageNet was used for feature extraction. The training set consists of 70,000 randomly sampled images, and the test set includes the remaining 30,000 images. Table \ref{table:table4} shows that ESH1 and ESH2 achieved the best performance in terms of mAP for 16, 32, 64, and 128 bits. For precision@1000, DSH achieved the best result, and ESH2 achieved a 2\% lower precision. However, as the number of bits increases, the performance of the ESH algorithms increases with a stronger trend such that for 32-bit settings ESH2 and DSH perform equally, and ESH algorithms outperform DSH for 64 and 128 bits with a margin of 3\% and 5\%, respectively. Clearly, the ESH algorithms attain the best precision at @r=2 under the 64-bit setting. Although for 16-and 32-bit settings, DSH and LGHSR achieved the best performance, the ESH algorithms were still competitive. The third row in Fig. \ref{fig:1}, illustrates how the ESH algorithms perform compared with recent methods in terms of precision-recall graphs. In almost all cases, the ESH algorithms achieved a better performance in comparison with the other methods.

\subsection{Effect of Regularization}

To determine how incorporating the regularization term in Eq.\ref{eq:7} improves the quality of the binary codes, we ran experiments with $\alpha = 0$ on NCT-CRC-HE-100K. Table \ref{table:table5} shows that the setting of $\alpha=0$ significantly degrades the performance, confirming the effectiveness of the proposed regularization.
\begin{table}[htb]
\centering
\setlength{\tabcolsep}{4pt}
\caption{Effect of the regularization term on retrieval performance in terms of mAP for NCT-CRC-HE-100K detaset.}
\begin{tabular}{|l|cccc|}
\hline
     
                  & \multicolumn{4}{c|}{NCT-CRC-HE-100K}                                                                                                              \\ \hline 
Method, Regularization 
                    & \multicolumn{1}{c|}{16 Bit}       & \multicolumn{1}{c|}{32 Bit}       & \multicolumn{1}{c|}{64 Bit}     & \multicolumn{1}{c|}{128 Bit} 
                   \\
                   
                   \hline\hline                            
ESH1, $\alpha=0 $                      & 40.65                               &     35.96                   &31.09       
&26.80                                     \\ \hline
ESH1, $\alpha$=automatic                       &66.32	  &66.77	&67.14	&66.26                                                           \\ \hline
\end{tabular}
\label{table:table5}
\end{table}

\subsection{Time Complexity and Runtime Comparison}
\begin{figure}[h]%
    \centering
\includegraphics[width=.5\columnwidth]{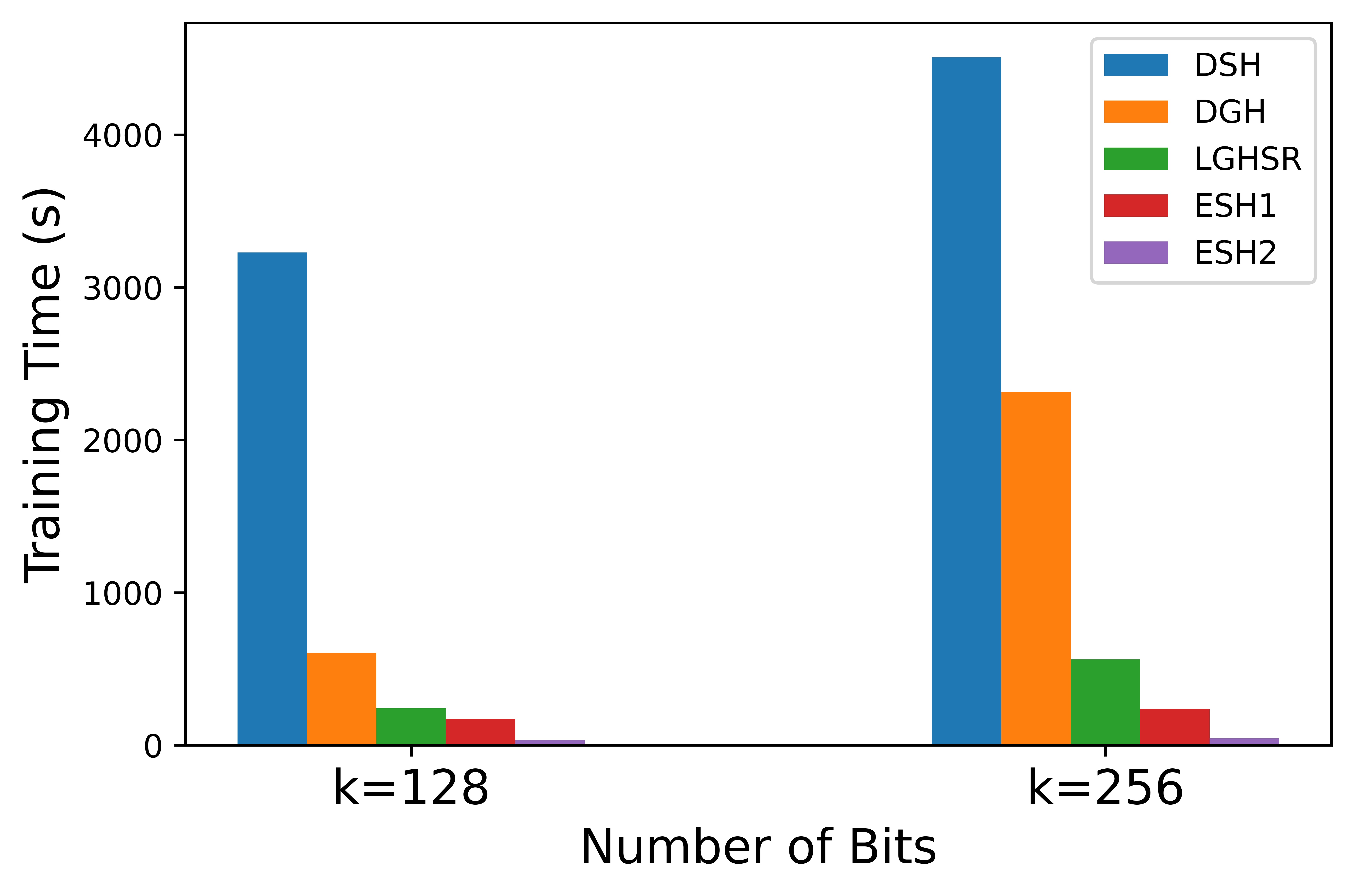}\\
\caption{Runtime comparison of ESH1 and ESH2 against the graph hashing algorithms DSH, DGH, and LGHSR on NUS-WIDE dataset for 128- and 256-bit settings.}%
    \label{fig:2}%
\end{figure}

\begin{figure}[h]%
    \centering
\includegraphics[width=.5\columnwidth]{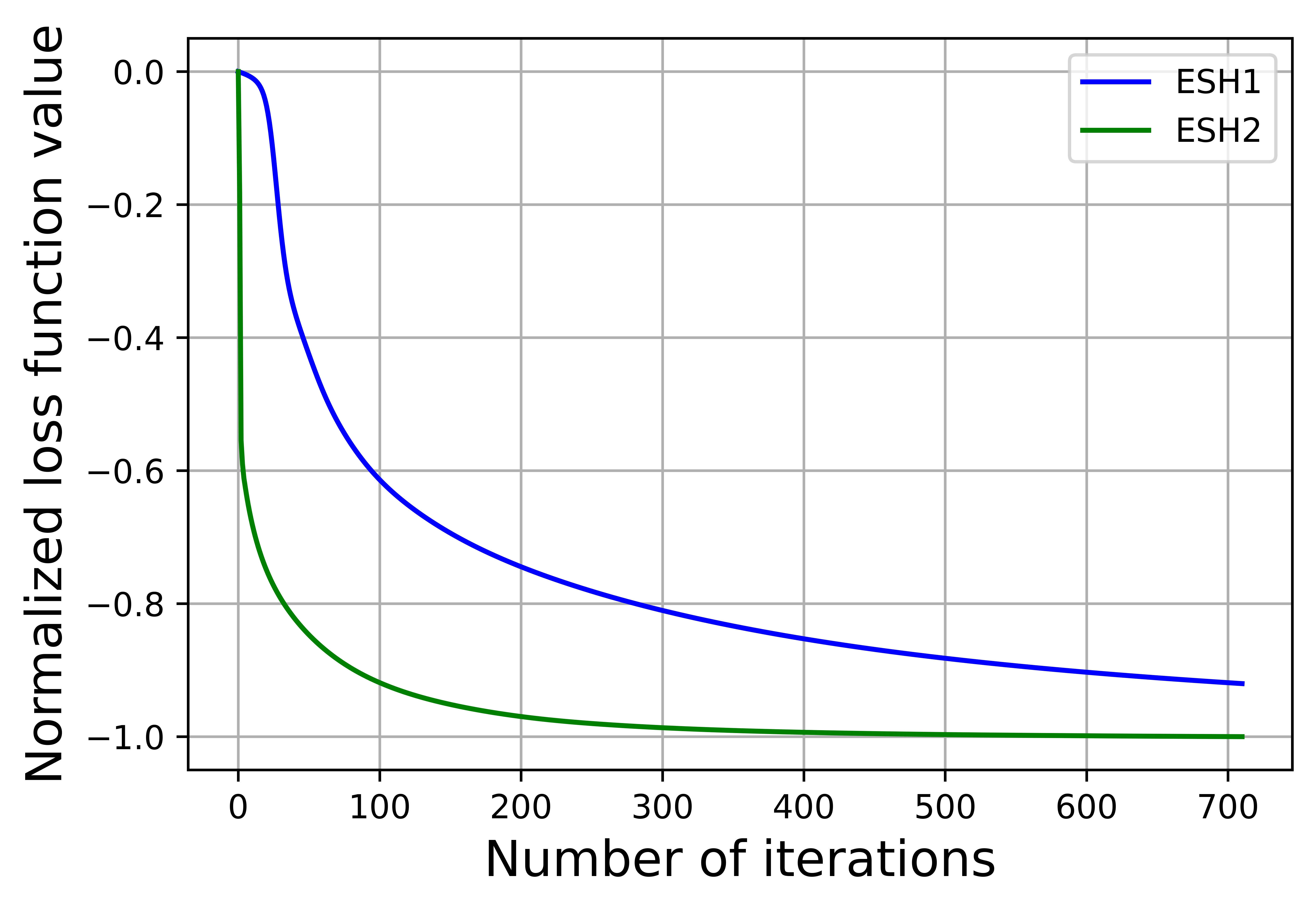}\\
\caption{Convergence comparison of ESH1 and ESH2 on NUS-WIDE dataset for 128-bit setting.}%
    \label{fig:3}%
\end{figure}

In this section, we provide a complexity analysis of the ESH1 algorithm and compare it with recent representative graph hashing methods. The complexity required for calculating matrix $\mathbf{S}$, which only needs to be calculated once, is $O(ndm)$. Note that the complexity of k-means algorithm for selecting the centers is excluded as this is a common step in all graph hashing algorithms which employ the low-rank approximation method proposed in AGH paper for affinity matrix construction. For the learning rule applied in the projected gradient method, the complexity is $O(2ndkN)$, and as a result, the overall complexity is $O(2ndkN+ndm)$. Table \ref{table:table6} presents the complexity of the proposed ESH algorithms and recent representative graph hashing methods without including the complexity of calculating the affinity matrix. Clearly, the RDSH algorithm has the highest time complexity $O(n^3)$, which is due to solving a
standard Sylvester equation to derive the spectral solution. In contrast, the AGH algorithm has the lowest complexity because it uses an eigen decomposition to obtain a spectral solution. The other algorithms listed in Table \ref{table:table6} are a function of at least two decision variables, which increases the complexity. For example, the term $nkNlog_2n$ in LGHSR and DSH represents the complexity of updating additional variables non-existent in the ESH algorithms. On the other hand, ESH has lower complexity as the optimization is a function of one variable, and there is no inner loop, for example, terms such as $N_GN$ in DSH or $N_BN$ in DGH iterations.

To further validate the training time efficiency of the proposed ESH algorithms, we conduct a runtime comparison among graph hashing methods on NUS-WIDE dataset for 128 and 256-bit settings. As discussed earlier, due to the closed-form solution for AGH, this algorithm has the lowest time complexity. By contrast, the time complexity of RDSH is $O(n^3)$, which is significantly higher than that of the other algorithms. As a result, we compared the runtime of the ESH methods with DSH, DGH, and LGHSR.
Fig. \ref{fig:2} shows the runtime of ESH1, ESH2, DSH, DGH, and LGHSR. Apparently, ESH2 and ESH1 have the lowest runtimes compared with the other graph hashing methods. In addition, based on this figure, the runtime of DGH and LGHSR increases quadratically with the number of bits i.e., $k$. This can also be seen in Table \ref{table:table6} where the $k^2$ term exists in the time complexity of DGH and LGHSR.

\subsection{ESH1 vs ESH2:}

As indicated in the results section, ESH2 (manifold optimization) achieves a better performance in most cases compared with ESH1 (projected gradient). This is because, in ESH2, the matrix $\mathbf{W}$ is updated on the Stiefel manifold. In other words, in ESH2, while $\mathbf{W}$ is updated, the orthogonality constraint is preserved. However, in ESH1, we update $\mathbf{W}$ using a gradient descent without considering the orthogonality constraint, and then find the closest orthogonal matrix to the updated $\mathbf{W}$. ESH2 and ESH1 can be compared in terms of the run time. As shown in Fig. \ref{fig:2}, ESH2 has a lower training time compared with ESH1, which is due to the faster convergence of the manifold optimization. The graph in Fig. \ref{fig:3} compares the convergence of ESH1 and ESH2 for the NUS-WIDE dataset. Clearly, ESH2 has a faster convergence rate. Note that although ESH2 performs better in terms of both the quality of the binary codes and the runtime, it may face a memory bottleneck during training owing to the matrix inversion step (see Eq. \ref{eq:14}). However, in our case, this is not a problem because the inversion is conducted on a $d \times d$ matrix instead of an $n \times n$ matrix where $d\ll n$. This is due to the fact that we transformed the problem from $n \times k$ parameters to a problem with $d \times k$ decision variables.

\begin{table}[hbt]
\centering
\setlength{\tabcolsep}{4pt}
\caption{Comparison of time complexities.}

\begin{tabular}{|l|c|}
\hline

Method     
& \multicolumn{1}{c|}{Time complexity }       
                                                                                             \\ \hline \hline

                    \hline                          
AGH                  & $O(m^2n+(s+1)kn)$                             \\ \hline
DGH                   & $O(nmkN_BN+k^2nN)$ 
 \\ \hline
RDSH         &$O(n^3)$                                \\ \hline
 
LGHSR        &$O(m^2n+(s+1)kn+2nk^2N+nkN \log_2n)$
\\ \hline

DSH   &$O(nmkN_GN+nkN \log_2n)$
\\ \hline

ESH   &$O(2ndkN+ndm)$
\\ \hline

\end{tabular}

\label{table:table6}
\end{table}

\subsection{Comparison with deep unsupervised hashing}

Although deep learning has mainly been applied to supervised hashing problems, recent attempts have been made to develop deep unsupervised hashing methods. Some examples include DH \cite{erin2015deep}, UHBDNN \cite{do2016learning}, DeepBit \cite{lin2016learning}, and SADH \cite{shen2018unsupervised}.
Table \ref{table:table7} compares the performance of our proposed method compared with these algorithms. Clearly, based on Table \ref{table:table7}, ESH 1 and ESH 2 outperform UHBDNN and DeepBit and provide competitive results compared with SADH. In this Table, \say{R+} implies that raw images are fed to the network, suggesting that these methods are end to end. By contrast, \say{V+} indicates that the vector representations are fed to the network.Considering the simplicity of our proposed method in comparison with deep learning methods, and that feature learning has not been included, the obtained results are promising

\begin{table}[!htb]
\centering
\small
\caption{ESH compared with deep unsupervised hashing algorithms for NUS-WIDE dataset. The R+ and V+ means the respective algorithm works on raw images and vector data (images after feature extraction) respectively. 
}
\begin{tabular}{|l|ccc|ccc|}
\hline
                  & \multicolumn{3}{c|}{mAP \%}    & \multicolumn{3}{c|}{precision \% @5000}                                                                                                                                                                      \\ \hline
Method                    & \multicolumn{1}{c|}{16 B}       & \multicolumn{1}{c|}{32 B}       & \multicolumn{1}{c|}{64 B}      & \multicolumn{1}{c|}{16 B}  & \multicolumn{1}{c|}{32 B}       & \multicolumn{1}{c|}{64 B}             \\ \hline\hline  
V+UHBDN               & 54.26                                  & 51.72                              & 54.74                              &70.18     &69.60        &72.74                                            
\\ \hline
 R+DeepBit                     &39.22                                  & 40.32                              &   42.06                             & 45.54    & 51.34   & 57.72                                                       \\ \hline
 R+SADH                    &\textbf{60.14}                                   &\textbf{57.99}                                &56.33                                 & \textbf{71.45}    & \textbf{73.88}    & 75.04                                                 \\ \hline

ESH1            &56.32  	 &56.89	   &57.47	 	   &67.58	  &69.47	 &71.46	                   \\ \hline

ESH2            &56.54   &57.16    &\textbf{57.71}   &69.74   &72.49     &\textbf{75.62} 

              \\ \hline
\end{tabular}
\label{table:table7}
\end{table}

\section{Conclusions}
In this paper, we proposed a novel formulation for spectral hashing that achieves a highly competitive performance compared with most recent methods, and at the same time achieves a low complexity. The proposed projected gradient method is highly efficient for three reasons. First, the formulation for ESH transforms the decision variable with a dimensionality of $n \times k$ into $d \times k$, where $d\ll n$. Second, the affinity matrix, which is $n \times n$ in the spectral formulation, was removed, and instead, a $d \times d$ matrix $\mathbf{S}$ plays a similar role. Finally, and more importantly, unlike other graph hashing schemes, the proposed formulation achieves high-quality binary codes without adding any additional decision variables to the problem. We applied two different optimization techniques, that is, a projected gradient and manifold optimization, to obtain a solution. Using extensive experiments on four public datasets, we showed that the proposed method outperforms or achieves highly competitive results compared with recent methods and offers a low complexity at the same time. For future work, we plan to update the affinity matrix along with the proposed loss function, which needs the use of an end-to-end training framework where feature learning is achieved through training. To conduct feature learning, we may need to employ a reconstruction loss with the proposed loss function in Eq. \ref{eq:7}. We believe that such a scheme can significantly improve the performance of the proposed method. Furthermore, another interesting path for future studies is to apply this more efficient non-alternating hashing scheme instead of the more complex alternating algorithm employed by \cite{hu2018hashing} for the network quantization problem.

\bibliographystyle{unsrt}  
\bibliography{ms}  
\end{document}